\documentclass[%
 reprint,
 amsmath,amssymb,
 aps,
]{revtex4-2}
\usepackage{xcolor}
\usepackage{amsmath}
\usepackage{graphicx}% Include figure files
\usepackage{dcolumn}% Align table columns on decimal point
\usepackage{bm}% bold math
\begin{document}
\setlength{\textfloatsep}{10pt plus 1.0pt minus 2.0pt}
\preprint{APS/123-QED}

\title{Learning grammar with a divide-and-concur neural network}

\author{Sean Deyo}
\email{sjd257@cornell.edu}
\affiliation{Cornell University}
\author{Veit Elser}
\affiliation{Cornell University}

\date{\today}

\begin{abstract}
 We implement a divide-and-concur iterative projection approach to context-free grammar inference. Unlike most state-of-the-art models of natural language processing, our method requires a relatively small number of discrete parameters, making the inferred grammar directly interpretable --- one can read off from a solution how to construct grammatically valid sentences. Another advantage of our approach is the ability to infer meaningful grammatical rules from just a few sentences, compared to the hundreds of gigabytes of training data many other models employ. We demonstrate several ways of applying our approach: classifying words and inferring a grammar from scratch, taking an existing grammar and refining its categories and rules, and taking an existing grammar and expanding its lexicon as it encounters new words in new data.
\end{abstract}

\maketitle

\section{Introduction}\label{sec:introduction}
Children display an innate facility for acquiring language. Beginning with a small vocabulary of word fragments, most humans are eventually able to grasp the meaning of arbitrarily long and convoluted strings of words automatically, even if the effort is not rewarded until the very end of the sentence. How is this ability gleaned from the sparse data children are presented with, a training corpus that comes nowhere close to sampling the full expressive power of language?

Language has \textit{syntax rules} that are acquired long before they are understood consciously in an instruction setting. Readers of this journal would accept ``Left-handed heterodyne detection of entangled meso-phase supernovas" as a grammatically valid title, even while questioning its scientific legitimacy. Humans seem to be able to grasp most elements of grammar without ever being told about nouns, verbs, etc.

Separate from the process of syntax-rule acquisition is the very question of what constitutes the right or cognitively most relevant set of rules. Human linguists have struggled with this question for over two centuries and have arrived at solutions (with several variations) for many natural languages. Could there be significantly different solutions that also ``explain" the data?

This study was motivated by all of the questions above and the desire to study them objectively. Can grammar be acquired without formal instruction, that is, in an unsupervised learning setting? Can the learning be implemented in a distributed manner, say on a network? Can the learning of abstract rules be demonstrated, that is, not just the production of language that is consistent with such rules? And if successful, how does the acquired grammar compare with the grammars developed by human linguists?

We make some concessions in addressing these research objectives. First, our model for representing grammar is not completely open-ended, but is based on the context-free grammar (CFG) model \cite{jurafsky} already introduced by linguists. However, CFGs are very general and also arise outside of natural language modeling. In our use of this model the categories are abstract entities that only acquire interpretations as ``parts of speech" during training. Second, in order to rigorously test our learning model we train on data generated by explicit model grammars rather than natural language data. This work should be seen as a proof-of-concept exercise. We make no claims that our particular model and its implementation on a network bear any strong relationship to reality (neuroethology).

Current-day natural language processing (NLP) methods achieve high scores in imitating language by accessing very large network-parameterized representations distilled from even larger collections of training data. Parameter sets and training corpora measuring in the terabytes are becoming commonplace \cite{brown2020language}. This approach is sometimes criticized as simple mimicry, and that the high fidelity in language production comes without any understanding \cite{marcus2018deepest,bender21on}. While the systems we train also do not understand meaning (semantics), we \textit{can} claim that they at least understand the syntax rules of the abstract entities in the grammar. This much of language is given a fully transparent, interpretable representation in our approach. Moreover, we find that this part of language learning is possible without the terabytes of data used in current NLP.

\section{Comparison with previous work}\label{sec:prevwork}
State-of-the-art language models (LMs) such as GPT-3 \cite{brown2020language}, trained mostly without supervision, would appear to have already solved the grammar inference problem in that the language they generate has very high grammatical accuracy. However, a linguist might argue that to demonstrate grammar understanding, one should also be capable of generating grammatically correct but semantically nonsensical output, something which is beyond models of this kind. The internal representation of grammar in these systems is inextricably linked with particular extracts of the training data. While the difference between the linguist's abstract and these example-based representations may not matter for some applications, it is surely relevant when modeling language acquisition and processing in humans. In this respect our approach, which uses abstract categories, is closer to the linguist's concept of grammar inference.

The representation of grammar in our approach falls into the connectionist paradigm but also differs in significant ways from current practice. Starting with Elman's simple recurrent networks (SRNs) \cite{elman1990finding}, the time structure of language has motivated designs that try to capture phrase structure, subject-object relationships, relative clauses, etc. in networks that take sequential data. In transformer networks \cite{vaswani2017attention}, the most sophisticated connectionist machines of this type, broad contexts for tokens in the stream are provided by an attention mechanism. In our approach, representations are distributed as well, but without an explicit reference to time. Instead, the network instantiations represent the parse trees of whole sentences, and have a direct linguistic interpretation.

In addition to treating grammar as an independent learned component of language, our approach differs from most current NLP research in its core technology. The encoders and decoders of transformer models are built with feed-forward neural networks that form representations of tokens (strings of words) in a continuous Euclidean space. Continuity of the representation space is required because the optimization performed in training is based on gradient information. Our representations live in Euclidean space as well, but for a different reason. The elementary operations are not gradient steps but distance-minimizing ``projections" to the nearest element of a set \cite{elser2021learning}. The latter can be discrete, where they represent symbolic entities such as categories and rules. For example, when representing categories by $k$-tuples of real numbers, by the usual 1-hot encoding, projection to a category takes the form of replacing the largest element with a 1 and setting the rest to 0, as that minimizes the distance to the constraint set.

Besides using discrete points and projections to them when processing the grammatical content of a sentence (categories, parse tree), we also encode the rules of the grammar discretely. In fact, our algorithm does not treat projections to nearest-category or nearest parse-tree any differently from projections to nearest rule-table. The discreteness of the rule table ``parameters" in our approach is the most obvious departure from standard practice in machine learning, and also key to bringing interpretability to the representation. 

{Large LMs often use hundreds of gigabytes of training data to fine-tune hundreds of billions of parameters~\cite{bender21on}. Having so many parameters makes it practically impossible for a human to interpret what the algorithm has learned. In the years before the current era of large LMs, some smaller models aimed to directly extract an interpretable set of grammatical rules from data, as we do. One of the best known is \mbox{\textsf{SEQUITUR}}~\cite{sequitir}, an algorithm which compresses a string of symbols into a set of context-free production rules based on recurring subsequences. Such an approach can effectively reproduce the data from a compact set of rules, but lacks the generative capacity to create novel sentences --- combinations of words not seen in the data that are nonetheless `grammatical' in the usual sense. The capacity for novelty often goes hand in hand with the ability to group words based on how they are used (i.e., identify parts of speech). Some models, such as \textsf{CDC}~\cite{CDC} and \textsf{ALLiS}~\cite{ALLiS}, take advantage of data that have already been tagged with grammatical labels on all of the words. Others, like us, demand that the algorithm learn the lexical categories without supervision (annotations). The \textsf{ADIOS} algorithm~\cite{ADIOS} constructs a graph whose vertices are words, encoding each sentence in the data set as a path through the graph. It discovers parts of speech by identifying high-probability sub-paths in the graph and creating equivalence classes of words that appear as parts of such patterns. The most significant pattern then becomes a new non-lexical symbol, which is added to the graph as a new vertex, and the process repeats. The \textsf{eGRIDS} algorithm~\cite{eGRIDS} also updates its grammar iteratively, starting with an initial hypothesis and merging or creating new nonterminal symbols with the goal of minimizing the complexity of the resulting grammar. The \textsf{CLL} approach~\cite{CLL} also proceeds stepwise, adding sentences one by one, updating the grammar to accommodate the newest sentence. Though these models share our aim of extracting interpretable grammatical rules from data, our method has a fundamentally different character. The existing methods are explicitly incremental, whether by a greedy search~\cite{ADIOS,CDC} or by stepwise updates to the grammar motivated by minimizing complexity~\cite{eGRIDS,CLL}. By contrast, our projection-based hard constraint approach attempts to solve the entire problem --- all of the sentences, all of the syntactic rules, all of the lexical categorizations --- simultaneously in a distributed framework.}

The projection approach to network optimization was introduced only recently \cite{elser2021learning}, which might explain why it is not more widely used. The competition between constraints makes it difficult to project to all of them simultaneously. The divide-and-concur technique \cite{gravel2008divide} resolves this difficulty by replicating variables so that all projections involve only easy constraints on small sets of independent variables (divide). Projecting to an equality constraint enforces agreement among the replicated variables (concur). In section \ref{sec:methods} we describe our projections and how we coordinate them to converge on solutions.

Finally, our method stands in stark contrast with current NLP practice in that it is possible to train on much smaller data sets, at least for the more limited task of learning grammar. Though the set of possible grammars expressible by our model is very large, the number of discrete parameters or bits of information to be learned is modest. The learning of cellular automata rules, like Conway's Game of Life, presented the same contrast \cite{elser2021reconstructing}. The discretely parameterized model needed only $2^n$ bits to represent the rule of an $n$-input automaton and could be trained on as few as a single pair of patterns, whereas the continuous model with gradient descent \cite{springer2020its} needed to be tenfold over-parameterized and used one million data. {Similarly, typical large LMs can have hundreds of billions of parameters and are trained on hundreds of gigabytes of data~\cite{bender21on}. Even the smaller LMs need hundreds~\cite{eGRIDS} if not thousands~\cite{ADIOS,CDC,CLL} of training sentences, especially if they are tasked with generating grammatical output as opposed to merely parsing existing sentences.} We will show that our algorithm only needs a handful of sentences in order to infer the rules of a simple grammar. %Much like a human, our algorithm can identify patterns even when data are scarce. 

\section{Problem statement}\label{sec:problem}

\subsection{Context-free grammar model}\label{sec:cfg}
A context-free grammar (CFG) is a formal grammar consisting of a \textit{lexicon} and a set of \textit{production rules}. The lexicon consists of \textit{terminals} --- the words appearing in the language --- and \textit{nonterminals} --- symbols that represent \textit{lexical categories} (noun, verb, etc.) and higher abstractions (noun phrase, verb phrase, and so on), called \textit{non-lexical categories}~\cite{jurafsky}.

\begin{figure*}[t]
  \centering
  \includegraphics[width=.8\textwidth]{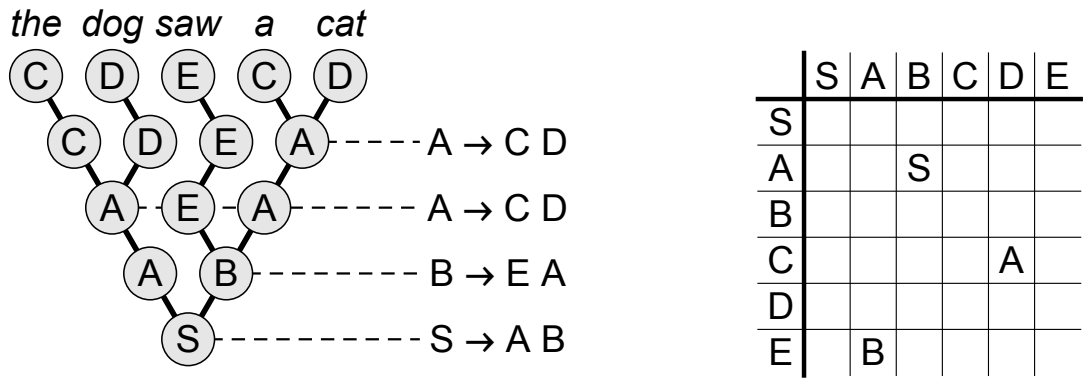}
  \caption{\textit{Left:} A parse tree for \textit{the dog saw a cat}. The words sitting on the top layer of the tree imply the lexical rules, such as \textsf{D} $\to$ \textit{dog}. \textit{Right:} The syntactic rules used in the parse tree, displayed as a binary operation table. The \textsf{S} in cell \textsf{AB} represents the rule \textsf{S} $\to$ \textsf{AB}, and so on for the other entries. In addition to having a linguistic interpretation, these diagrams are faithful depictions of the architecture used by our algorithm. Each node in the parse tree and every cell of the rule table holds a $c$-component category vector.}
  \label{fig:treeandrules}
\end{figure*}

Generally, the rules take the form \textsf{X} $\to$ $\alpha$ where \textsf{X} is a single nonterminal and $\alpha$ is a string of symbols (terminal or nonterminal). To simplify the structure of the grammar we allow only two kinds of rules:
\begin{itemize}
    \item \textit{Syntactic rules} of the form \textsf{X} $\to$ \textsf{YZ} where \textsf{X} and \textsf{Y} $\neq$ \textsf{Z} are nonterminals.
    \item \textit{Lexical rules} of the form \textsf{X} $\to$ $w$, where \textsf{X} is a nonterminal and $w$ is a terminal (word).
\end{itemize}
%We also require that there be exactly one lexical rule for each word. Thus, the lexical rules partition the set of words, and the set of lexical symbols is the set of non-terminals that appear in lexical rules.

One special symbol, \textsf{S}, serves as the ``start symbol." Any valid sentence must be derivable by taking a single \textsf{S} and repeatedly applying rules until only terminals remain~\cite{jurafsky}. Figure \ref{fig:treeandrules} displays this process with a \textit{parse tree}. Each branching event represents the use of a syntactic rule, and the words above the final layer of the tree imply the lexical rules. The fact that a rule replaces a single category, without reference to its neighbors, is what makes the grammar context-free.

The binary restriction on the syntactic rules gives the parse trees the triangular structure in Figure \ref{fig:treeandrules}. This structure simplifies the architecture of the networks that our inference algorithm will use. To represent ternary rules, such as \textsf{NP} $\to$ \textsf{NP} \textsf{AND} \textsf{NP} (conjunction of noun phrases), our restricted model would have to `invent' auxiliary categories that fit the binary restriction (with rules \textsf{NP} $\to$ \textsf{NPA} \textsf{NP} and \textsf{NPA} $\to$ \textsf{NP} \textsf{AND}). The further restriction that the \textsf{Y} and \textsf{Z} of the rule are distinct is reflected in natural language grammars but could be relaxed.

One can make further refinements within a category based on \textit{features}. Features express properties such as the number of a noun (singular or plural), the tense (past, present, etc.) of a verb or its mood (indicative, interrogative, etc.). We will denote features with a subscript: e.g., the rule $\textsf{NP}_\mathsf{s} \to \textsf{D}_\mathsf{s} \textsf{N}_\mathsf{s}$ represents the replacement of a singular noun phrase with a singular determiner followed by a singular noun. 

Before training, the algorithm does not ``know'' which words are singular or plural, nor does it ``know'' what a noun is, so we simply label the lexical and non-lexical categories with \textsf{A}, \textsf{B}, \textsf{C}, and so on; and the feature subscripts with $0$, $1$, and so on. Only after the algorithm finds a solution can one notice that, for example, \textsf{A} happens to contain all the nouns and $\mathsf{A}_0$ happens to have all the singular nouns.

\subsection{Restrictions and hyperparameters}\label{sec:restrictions}

When data are limited --- only a small number of sentences are available --- and the CFG model is unrestricted, we should not expect the grammar inference problem to have a unique solution. Sentences generated by the inferred grammars will almost always be ungrammatical within the language from which the data was sampled. To promote unique grammar inference, even in this data-limited setting, we next introduce some restrictions in the form of hyperparameters. Note that the restrictions in the definition of the syntactic rules of section \ref{sec:cfg} are technical in nature and do not address the uniqueness question.

Our approach to promoting unique inference is based on a max-min principle involving two hyperparameters: the number of lexical categories $c_l$ and the number of syntactic rules $r_s$. Clearly one would like to be able to resolve the constituents of sentences to the greatest extent possible --- maximizing $c_l$ --- while at the same time using the fewest rules --- minimizing $r_s$ --- in the parse trees that generated them.

We implement the maximizing principle by making $c_l$ a hyperparameter and imposing the constraint that all $c_l$ lexical categories appear in the top layers of the parse trees. If $\Omega$ is the set of distinct words in the data, then the grammar needs $|\Omega|$ lexical rules that map the words surjectively to the $c_l$ lexical categories. We allow for the possibility of homographs (words with the same spelling but different meaning) by allowing the number of lexical rules to exceed $|\Omega|$ by another hyperparameter $h\ge 0$.

Our hyperparameter-imposed restrictions are summarized as follows:
\begin{enumerate}
\item All $c$ categories, of which at least $c_l$ are lexical, must be used.

\item There are at most $r_s$ syntactic rules.

\item There are at most $r_l=|\Omega|+h$ lexical rules.
\end{enumerate}

A simple protocol for the max-min optimization is to set $c_l$, $r_s$ and some homograph allowance $h\ge 0$. If a solution is found, one increases $c_l$, decreases $r_s$, or decreases $h$ until solutions are no longer found. Minimizing the number of non-lexical categories might achieve the same end as minimizing $r_s$. In practice we set the total number of categories $c$, of which $c_l$ are reserved to be lexical, and maximize $c_l$. In section \ref{sec:results} we give examples of settings of all these hyperparameters that yield unique grammar inference.

Let $\Sigma$ be the set of nonterminals. We also impose the following restrictions that refer to this set:
\begin{enumerate}
\setcounter{enumi}{3}
\item The start symbol \textsf{S} must appear at the base of every tree, and nowhere else in the tree.

\item The mapping from $\Sigma$ into $\Sigma\times\Sigma$ defined by the rule set is injective; that is, if \textsf{U} $\to$ \textsf{XY} and \textsf{V} $\to$ \textsf{XY} are rules then $\mathsf{U}=\mathsf{V}$.

\item In any rule \textsf{X} $\to$ \textsf{YZ}, the \textsf{X} cannot be one of the $c_l$ designated lexical categories. If \textsf{X} $=$ \textsf{S}, then \textsf{Y} and \textsf{Z} cannot come from the $c_l$ designated lexical categories either.

\end{enumerate}

The injectivity constraint is motivated by the idea that the process of contracting a sentence down the layers of the parse tree should have something to do with the extraction of the sentence's meaning. Making the grammatical contractions deterministic presents the extraction of meaning with fewer choices, which makes the meaning less ambiguous.

Even with these restrictions, deriving a single sentence in isolation is usually trivial, especially if it contains no repeat words. The real work of uncovering the patterns of a grammar takes place when the set of sentences is large enough for most or at least some of the words to appear multiple times. For example, the solution in Figure \ref{fig:treeandrules} takes \textit{the} to be of category \textsf{C}. If there are other sentences in the dataset containing \textit{the}, it must always appear as category \textsf{C}. The algorithm might then recognize that any word following \textit{the} is likely be of category \textsf{D}, and thus the algorithm learns how to classify nouns.

\section{Algorithm}\label{sec:methods}
Our algorithm involves two variable types:
\begin{itemize}
    \item The \textit{category vectors}, $v$, are a collection of one-hot vectors representing which category is present at each node of each parse tree: If $v^{s\ell n}_\mathsf{A}=1$, then category \textsf{A} is present at node $n$ of layer $\ell$ of the parse tree for sentence $s$.
    \item The \textit{syntax tensor}, $t$, encodes the syntactic rules of the grammar: If $t_\mathsf{ABC}=1$ then \textsf{A} $\to$ \textsf{BC} is one of the syntactic rules.
\end{itemize}
The restrictions on the CFG described in section \ref{sec:problem} were in part motivated by keeping our network architecture simple. In particular, by having only binary syntactic rules the parse trees can be represented by a fixed set of category vectors arranged in a triangle as in Figure \ref{fig:treeandrules}. The same restriction allows us to represent the syntactic rule set as a third-order tensor. By the injectivity restriction there can be at most a single $1$ over the first index when the other two are fixed. In a solution, where this constraint is satisfied, the syntax tensor can be displayed as a binary operation table as in Figure \ref{fig:treeandrules}. The bits in the syntax tensor roughly correspond to the ``switches" in Chomsky's universal grammar \cite{chomsky2000new}.  

Expressed in terms of our two variable types, the task of the algorithm is to populate the syntax tensor with $1$'s such that there is a compatible assignment of category vectors to all of the trees. The algorithm itself is based on a ``divide and concur'' approach. The key is having multiple copies of the variables --- a new copy for every action in which the variable is involved. For instance, a sentence with five words requires four syntactic rule applications to get from one start symbol to five lexical categories, so there are four copies of the syntax tensor. We use $t^{s\ell}$ to denote the copy of the syntax tensor used at layer $\ell$ of sentence $s$. Each category vector is used twice: once in connecting to the layer above, either by use of a rule or by preservation of a category from one layer to the next, and once in similarly connecting to the layer below. We therefore have two copies of the category vector at every node, except the nodes in the bottom and top layers of the trees, which only need one copy. We use $v^{s\ell n\uparrow}$ and $v^{s\ell n\downarrow}$ to denote the upward- and downward-facing copies of the category vector at node $n$ of layer $\ell$ of sentence $s$.

All of these copies allow us to divide the difficult global problem of explaining the entire data set into a collection of simple local problems: Making sure the $\downarrow$ category vectors in one layer can be obtained from the $\uparrow$ category vectors in the layer below by applying exactly one syntactic rule from the local copy of the syntax tensor. {Figure \ref{fig:Atree} gives an example of a parse tree in which each layer makes sense in isolation, but some nodes have disagreement between their two category vector copies. Having two copies of the category vectors at each node (except nodes in the top and bottom layers) and a separate copy of the syntax tensor for every layer is what makes it possible to handle each layer independently. For instance, in layer 2, \textsf{B C} expands to \textsf{C D C} using the rule \textsf{B} $\to$ \textsf{C D}. The copy of the rule tensor for that layer (not shown in the figure) must have $t_{\textsf{BCD}}=1$. The copies of the rule tensor in different layers may disagree with each other, and with the copies used in the parse trees of other sentences. To rectify this, after solving the local problems layer by layer and sentence by sentence, we enforce a separate constraint to make the local copies of the variables concur.

Here we give a high-level overview of the algorithm; appendices \ref{sec:PA} and \ref{sec:PB} describe the details of the projections, while appendix \ref{sec:DC} compares and contrasts divide-and-concur networks with feed-forward networks. Let us use $x=(v,t)$ as a shorthand to denote the state of all the copies of all the $v$ and $t$ variables. For a solution, these variables must be discrete (0's and 1's), but during the search we allow them all to be real numbers.} Let $A$ denote the set of $x$ that satisfy all the local problems --- that is, all $v$ and $t$ variables are $0$'s and $1$'s and each layer of category vectors can be obtained from the layer below by applying a syntactic rule that has a $1$ in the local copy of the syntax tensor. Let $B$ be the set of $x$ that make all the copies agree --- that is, all copies of the syntax tensor agree, the $\uparrow$ and $\downarrow$ category vectors agree at each node, and for each word the corresponding top-layer category vectors all agree. {Any point in $A\cap B$ is a solution: $A$ ensures the variables are discrete and make sense locally, while $B$ ensures all the copies agree. 

\begin{figure}[t]
  \centering
  \includegraphics[width=.45\textwidth]{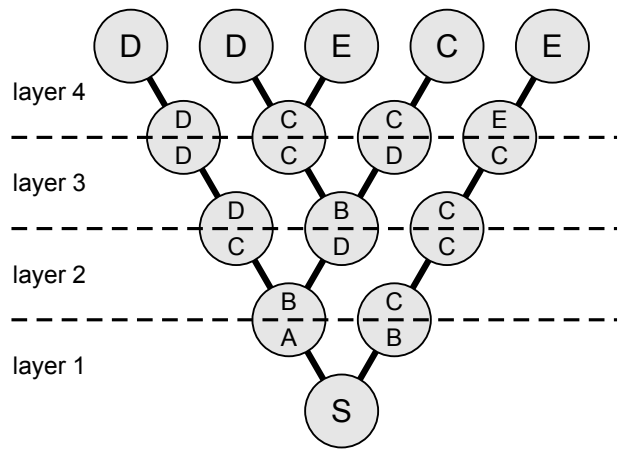}
  \caption{An example of a parse tree during the search process (before finding a solution), after projection to set $A$. Each layer involves the use of a single syntactic rule. For example, in layer 3 the \textsf{B} in the middle is replaced with \textsf{C D} and the other categories are passed along unchanged. Most of the nodes are involved in two layers, so they have two copies of their category vector, and the two copies do not always agree. Finding a collection of parse trees (one for each sentence) in which the copies do agree is part of the challenge.}
  \label{fig:Atree}
\end{figure}

Each iteration of the algorithm begins with $x$ as a vector of real numbers. We first find $P_A(x)$, the projection of $x$ to the nearest point in $A$. See appendix A for the details of this projection. We then compute the $A$ reflection: ${R_A(x)=2P_A(x)-x}$. 

The $B$ projection $P_B$ and reflection $R_B$ are defined analogously. Unlike $A$, $B$ does not require the variables to be discrete, so to compute $P_B$ we simply average the two copies of the category vector at each node, average all copies of the rule tensor, and so on (see appendix B for details).

The algorithm averages $x$ with its double reflection,
\begin{equation}
    x \mapsto x'=(1-\beta/2) x + (\beta/2) R_B(R_A(x)),
    \label{eq:RRR}
\end{equation}
where $\beta\in(0,2)$, and iterates until it converges on a point $x^*$ such that $R_B(R_A(x^*))=x^*$. We find that $\beta=0.5$ works well. One can see that if we succeed in finding such a point $x^*$, then $P_A(x^*)$ is indeed in $A\cap B$.}
%Let $P_i(x)$ be the projection to set $i$ and let ${R_i(x)=2P_i(x)-x}$ be the corresponding reflection, for $i=A,B$. Note that set $B$ does not require the variables to be discrete, so the projection is a simple average of the participating copies. A solution is any point in $A\cap B$: $A$ ensures the variables are discrete and make sense locally, while $B$ ensures all the copies agree. The algorithm initializes all the copies of the category vectors and syntax tensor to random values between $0$ and $1$ and iterates by averaging $x$ with its double reflection,
%\begin{equation}
%    x \mapsto x'=(1-\beta/2) x + (\beta/2) R_B(R_A(x)),
%    \label{eq:RRR}
%\end{equation}
%where $\beta\in(0,2)$, until it converges on a point $x^*$ such that $R_B(R_A(x^*))=x^*$. We find that $\beta=0.5$ works well. One can see that the point $P_A(x^*)$ is indeed in $A\cap B$, so that is a solution. 
We refer to the distance moved, $\|x'-x\|$, as the ``error" as this vanishes at a solution fixed point. More information about this ``relaxed-reflect-reflect" (RRR) algorithm can be found in \cite{elser2021learning}.

Making projections, of course, requires a choice of metric. The Euclidean metric is the default, but we need to modify it for the problem at hand. Since the $v$ and $t$ variables have fundamentally different roles, we allow them to have different weights in the metric. With negligible extra work in the implementation, we refine the metric further across the components of the category vectors:
\begin{equation}\label{metric}
    d((v,t),(v',t'))=\left(\sum_{\mathsf{X}\in\Sigma}\mu_\mathsf{X}^2\|v_\mathsf{X}-v_\mathsf{X}'\|^2\right)+\|t-t'\|^2
\end{equation}
where $\|\cdot\|$ is the standard L2 norm, and the metric parameters $\mu_\mathsf{X}>0$ express the relative weights for each category. We follow the practice described in \cite{deyo2021avoiding} for updating metric parameters adiabatically during the search. The purpose of updating the metric parameters in this fashion is to help avoid situations in which the algorithm gets stuck in a limit cycle or has some variable types fixed while the others wander fruitlessly.

There are two useful supplemental algorithms that we include in this work. First is the \textit{category refiner}. The idea of the refiner is to take an existing solution and refine its categories and rules further. For instance, one can use the main algorithm to work out the basic grammatical divisions --- separating nouns, verbs, etc., and learning how they relate syntactically --- then use the refiner to break the categories down into singular and plural forms, or perhaps masculine and feminine forms. One could try to capture these features from the beginning with the main algorithm by specifying larger values for $c$ and $c_l$, but we find that it is often quicker to run the main algorithm with modest $c$ and $c_l$ and then pass the solution to the refiner.

%We give full details in Appendix \ref{sec:refinerdetails}, but here we give an overview of the refiner algorithm. 
To use the refiner, one provides a set of syntactic and lexical rules and a collection of parse trees for the solved sentences. Rather than specifiying the number of categories, one specifies the number of features, say $f_\mathsf{A}$, into which category \textsf{A} is to be subdivided, and so on. With the parse trees in hand, the algorithm already knows which category is present at each node. What remains is to identify what feature of that category should be present: e.g., if the category is a noun, is it singular or plural? In practice this means placing a feature vector of length $f_\mathsf{A}$ at each node with category $\mathsf{A}$, and so on for the other categories. 

Refining features means that there are $f_\mathsf{X} \times f_\mathsf{Y} \times f_\mathsf{Z}$ possible refined versions of each syntactic rule \textsf{X} $\to$ \textsf{YZ}. Thus, instead of a single $c\times c\times c$ tensor we have $r_s$ tensors --- one for each (unrefined) syntactic rule --- each of which is $f_\mathsf{X} \times f_\mathsf{Y} \times f_\mathsf{Z}$. Rather than specifying the number of (unrefined) syntactic rules $r_s$, one specifies the number of refined syntactic rules $f_\mathsf{XYZ}$ to be allowed for each unrefined rule \textsf{X} $\to$ \textsf{YZ}. 

Rather than giving a metric parameter to every feature of every category, we use a single metric parameter $\mu$ for the feature vectors. The metric parameter updating scheme is not as effective at saving the refiner from getting stuck as it is with the main algorithm. Even so, it provides a helpful diagnostic: When the refiner is stuck, $\mu$ wanders far from unity. Whenever $\mu>10$ or $\mu<1/10$, we infer that the refiner is stuck and reset all the variables to random initial conditions and set $\mu=1$.

Apart from these changes the refiner proceeds in much the same way as the main algorithm, except that everything that used to represent a category now represents a feature. Once the refiner finds a solution, it can output a new set of rules and parse trees, which can be refined further if desired.

The second supplemental algorithm is the \textit{lexical extender}. The idea of the extender is to take an existing (syntactic and lexical) rule set and check if it can explain a new list of sentences, with the possible addition of more lexical rules if the new sentences contain words that are not present in the existing solution. As far as implementation, the extender is essentially the same as the main algorithm but with some of the variables fixed --- namely, all copies of the syntax tensor and any top-layer category vectors for which the corresponding word is already in the pre-extended lexicon.

\section{Experiments}\label{sec:results}
We now apply our algorithms to a few \textit{language fragments}: small toy grammars that contain only a small subset of the possible sentences in the complete language. The examples we use are from the Natural Language Toolkit (NLTK)~\cite{NLTK,loper2002nltk}. 

\begin{table}[t]
    \begin{tabular}{l}
        \textsf{S} $\to$ \textsf{NP} \textsf{VP}\\
        \textsf{VP} $\to$ \textsf{V} \textsf{NP} $|$ \textsf{V} \textsf{NP} \textsf{PP} \\
        \textsf{PP} $\to$ \textsf{P} \textsf{NP} \\
        \textsf{V} $\to$ \textit{saw $|$ ate $|$ walked} \\
        \textsf{NP} $\to$ \textit{John $|$ Mary $|$ Bob} $|$ \textsf{D} \textsf{N} $|$ \textsf{D} \textsf{N} \textsf{PP} \\
        \textsf{D} $\to$ \textit{a $|$ an $|$ the $|$ my} \\
        \textsf{N} $\to$ \textit{man $|$ dog $|$ cat $|$ telescope $|$ park} \\
        \textsf{P} $\to$ \textit{in $|$ on $|$ by $|$ with} \\[3pt]
        \hline
        \\[-5pt]
        \textit{Bob ate my man.}\\
        \textit{John saw my man by a telescope.}\\
        \textit{Bob ate John.}\\
        \textit{a cat in John saw John.}\\
        \textit{the cat saw the man on John.}\\
        \textit{a cat with Bob saw a telescope.}\\
        \textit{Mary saw John.}\\
        \textit{Mary saw the telescope.}\\
        \textit{Bob saw the park by Bob.}\\
        \textit{John saw a man.}\\
    \end{tabular}
    \caption{\textit{Top}: The rules of the first NLTK \cite{loper2002nltk} grammar that we used to generate data. The vertical bar $|$ represents a disjunction: e.g., \textsf{VP} has the two rules \textsf{VP} $\to$ \textsf{V} \textsf{NP} and \textsf{VP} $\to$ \textsf{N} \textsf{NP} \textsf{PP}. These rules do not satisfy the binary restriction we impose on solutions, so the algorithm will have to find a slightly different set of rules that still explains the data. \textit{Bottom}: Ten randomly generated sentences from this grammar.}
    \label{tab:grammar1}
\end{table}

Table \ref{tab:grammar1} gives the syntactic and lexical rules of our first grammar. To generate the data we start with \textsf{S} and apply its rule to obtain \textsf{NP} \textsf{VP}. Then we randomly apply one of the available rules for \textsf{NP} and for \textsf{VP}, and then continue randomly applying rules for any remaining nonterminals until we have a string of terminals --- that is, a sentence. This grammar is capable of infinite nested loops of prepositional phrases, so we discard sentences with three or more such phrases. Not all of the syntactic rules in this grammar have the binary form \textsf{X} $\to$ \textsf{YZ} and our algorithm will have to deal with this as explained in section \ref{sec:cfg} by creating auxiliary categories, or by using the smaller number of categories more creatively.

\begin{table}[t]
    \begin{tabular}{l}
        \textsf{S} $\to$ \textsf{A} \textsf{B} $|$ \textsf{A} \textsf{C}\\
        \textsf{A} $\to$ \textsf{A} \textsf{D} $|$ \textsf{B} \textsf{A} $|$ \textsf{C} \textsf{A} $|$ \textsf{C} \textsf{D} $|$ \textsf{D} \textsf{A} $|$ \textsf{D} \textsf{B}\\
        \textsf{B} $\to$ \textit{my $|$ a $|$ in $|$ the}\\
        \textsf{C} $\to$ \textit{Bob $|$ man $|$ John $|$ telescope $|$ cat $|$ with $|$ Mary}\\
        \textsf{D} $\to$ \textit{ate $|$ saw $|$ by $|$ on $|$ park}\\[3pt]
        \hline
        \\[-5pt]
        \textit{my by ate a saw a.}\\
        \textit{by in John.}\\
        \textit{Mary Bob saw cat.}\\
        \textit{John saw by with.}\\
        \textit{cat saw in on in.}\\
    \end{tabular}
    \caption{\textit{Top}: The rules inferred by the algorithm, given ten random sentences from the first toy grammar and $c=5$, $c_l=3$, $r_s=8$, and $h=0$. \textit{Bottom}: Five randomly generated sentences from this inferred grammar.}
    \label{tab:badgrammar1}
\end{table}

The general approach to using the algorithm is to start with a small number of categories and a large number of syntactic rules. In this regime the algorithm finds solutions quite easily, but the solutions will not be unique and will generate ``nonsense", i.e. sentences inconsistent with the grammar that was used to generate the data. Table \ref{tab:badgrammar1} gives an example, using $c=5$, $c_l=3$, $r_s=8$, and $h=0$. The sentences generated by the inferred grammar certainly do not match the grammar used to generate the data. The inferred grammar is too large, in the sense that it is capable not only of generating the data sentences but also many other nonsense sentences. We want the smallest possible grammar that contains the data, and for the right choice of parameters the inferred grammar should be unique up to permutation symmetry. To infer a better grammar, we reduce $r_s$ until the algorithm can no longer find solutions. Then we increase $c$ or $c_l$ and repeat, starting with a large $r_s$ and reducing it until solutions are no longer found. At every step along the way, we keep track of the sizes of the inferred grammars and whether the solutions are unique.

\begin{table}[t]
    \centering
    \begin{tabular}{l}
        $\textsf{S}\,\to\,\textsf{B}\,\textsf{C}$\\
        $\textsf{A}\,\to\,\textsf{D}\,\textsf{C}$\\
        $\textsf{B}\,\to\,\textsf{C}\,\textsf{E}$\\
        $\textsf{C}\,\to\,\textsf{C}\,\textsf{A}\,|\,\textsf{G}\,\textsf{F}\,|$ \textit{Bob $|$ John $|$ Mary}\\
        $\textsf{D}\,\to$ \textit{by $|$ in $|$ on $|$ with}\\
        $\textsf{E}\,\to$ \textit{ate $|$ saw $|$ walked}\\
        $\textsf{F}\,\to$ \textit{man $|$ telescope $|$ cat $|$ park $|$ dog}\\
        $\textsf{G}\,\to$ \textit{my $|$ a $|$ the}
    \end{tabular}
    \caption{The rules inferred by the algorithm, given random sentences from the first toy grammar and $c=8$, $c_l=4$, $r_s=5$, and $h=0$.}
    \label{tab:grammar1sol}
\end{table}

For this data set, the right combination seems to be $c=8$, $c_l=4$, $r_s=5$. Table \ref{tab:grammar1sol} gives one of the resulting solutions. Note that even though $c_l=4$, there are actually five lexical categories. Our stipulation of $c_l$ only enforces a \textit{minimum} number of lexical categories. The remaining categories can be purely syntactic, like \textsf{A} and \textsf{B}, or purely lexical, or a combination of syntactic and lexical, like \textsf{C} is in this solution.

{When we use the solution in Table \ref{tab:grammar1sol} to randomly generate new sentences, we find that this grammar uses prepositional phrases following proper nouns somewhat more liberally than the original grammar from Table \ref{tab:grammar1}. The original grammar only uses a prepositional phrase after a proper noun if it comes after the verb. For example, the original grammar would never output \textit{a dog with Mary in the park saw Bob}, but our solution grammar can. Such constructions occur in about $20\%$ of randomly generated sentences.}
%, although this frequency can change if we modify the method of generation. For instance, we can take advantage of the contextual information in the inferred parse trees: When choosing which rule to use for \textsf{C} in \textit{a dog with} \textsf{C} \textit{in the park saw Bob}, we can check the parse trees to see which rules were used for \textsf{C} when it had a preposition immediately before and after it. We would find that \textsf{C} $\to$ \textit{Mary} was never invoked in such cases, so we would exclude it when deciding which rule to use. If we use this almost-context-free method of generation, the discrepancy between the inferred grammar and the original grammar disappears. Of course, the ability to generate novel constructions may be desirable in some cases, so abandoning the purely context-free scheme is not necessary.

\begin{figure}[t]
    \centering
    \includegraphics[width=.45\textwidth]{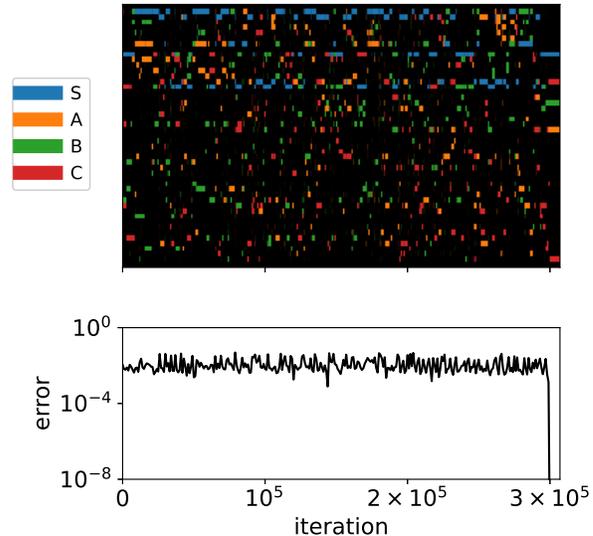}
    \caption{\textit{Top}: The evolution of the syntax tensor concur estimate, $P_B(t)$, as the algorithm searches for a solution to our first toy grammar dataset. One can think of each row as a cell in the syntactic rule table (e.g., the right side of Figure \ref{fig:treeandrules}). As the iteration count increases (moving from left to right), the changing colors reflect the algorithm's exploration of the many possible syntax tensors: If a row is black the corresponding cell of the rule table is empty; if not, the row's color indicates which category belongs in the corresponding cell. Even without the benefit of full color, one can appreciate the changing locations of the non-black splotches as the different cells of the rule table become active or inactive. \textit{Bottom}: Evolution of the error for the same run. The dramatic drop in error around $3\times10^5$ iterations signals that the algorithm has found a solution.}
    \label{fig:solving}
\end{figure}

Figure \ref{fig:solving} illustrates the dynamics of the algorithm as it searches for this solution. The upper panel represents the evolution of the concur estimate of the syntax tensor, $P_B(t)$. Each vertical slice represents the state of the tensor at a single moment, with the iteration number increasing from left to right. Each row corresponds to a pair of categories from \textsf{AA} at the top to \textsf{GG} at the bottom, and the colors indicate which (if any) rules are present. For instance, when the row for \textsf{BC} has a blue splotch, the algorithm thinks \textsf{S} $\to$ \textsf{BC} is one of the rules, but when the row is black, the algorithm does not think any rules of the form \textsf{X} $\to$ \textsf{BC} are present. (The rows for \textsf{AA}, \textsf{BB}, and so on are always black, since every rule \textsf{X} $\to$ \textsf{YZ} must have \textsf{Y} $\neq$ \textsf{Z}.) Categories \textsf{D} through \textsf{G} are designated as strictly lexical, so only \textsf{S}, \textsf{A}, \textsf{B}, and \textsf{C} can have syntactic rules. Even a reader without the benefit of full color can appreciate that most rows are black at any given moment, reflecting the sparsity of the syntax tensor, and that the non-black splotches change as the algorithm explores the space of possible syntax tensors. 

The lower panel of Figure \ref{fig:solving} gives the evolution of the rms error of all the variables for the same run. Just like the evolution of the syntax tensor, this time series is correlated over thousands of iterations, so that the actual number of solution candidates explored is much less than that implied by the iteration count. The error fluctuates around $10^{-2}$ for most of the run, before abruptly decreasing by several orders of magnitude after about $3\times 10^5$ iterations. This is the algorithm's ``aha moment'' when it discovers a solution. One can see in the upper panel that the syntax tensor remains fixed after this moment arrives.

\begin{table}[t]
    \centering
    \begin{tabular}{l|c|c}
        sentences & successes/trials & iterations/success\\
        \hline
        $10$ & $27/100$ & $(3.0\pm0.8)\times10^6$\\
        $20$ & $60/100$ & $(1.0\pm0.2)\times10^6$\\
        $50$ & $93/100$ & $(4.5\pm0.4)\times10^5$\\
        $100$ & $83/100$ & $(6.7\pm0.6)\times10^5$\\
    \end{tabular}
    \caption{Performance statistics for the algorithm on the first toy grammar. Each trial was limited to $10^6$ iterations.}
    \label{tab:grammar1stats}
\end{table}

Table \ref{tab:grammar1stats} summarizes the results of running $100$ trials of the algorithm with $c=8$, $c_l=4$, $r_s=5$, and $h=0$ (with random starts), for different numbers of sentences in the data. Each trial is limited to $10^6$ iterations. We update the metric parameters with rate $10^{-4}$ \cite{deyo2021avoiding}, meaning that a change in $\mu$ of order $1$ takes $10^4$ iterations. The final values of the $\mu$'s tend to be close to $0.7$ for non-lexical categories and $1.3$ for lexical categories, plus or minus a few tenths. One may wonder if the metric parameter updating is truly necessary, given that the final values are not far from unity. In fact it is: In $100$ trials on $10$ sentences without metric parameter updating, only one succeeded.

{With small data sets, many words might only appear once or twice, yet that can be enough to uniquely constrain the syntax. From $20$ sentences on, all of the inferred solutions were identical to the one in Table \ref{tab:grammar1sol} (up to permutation symmetries), and even with $10$ sentences that solution was the most common. In fact, we observed one sample of only $5$ sentences that always produced the solution in Table \ref{tab:grammar1sol}. Increasing the sentence count to $50$ or $100$ improves the success rate somewhat, perhaps because having more appearances for each word provides a stronger concur constraint, but still the solution is unchanged. A more complex language fragment may need more sentences to reach uniqueness, but at some point the extra sentences do not provide any new syntactic information --- just more parse trees to fill in, and perhaps more words to add to the lexicon.} It is worth noting that the extra data also makes each iteration take longer, as the algorithm must loop through every sentence's parse tree. One way to avoid this extra time is to have the algorithm work in batches, only looking at, say, $10$ sentences at a time. Another option would be to have the algorithm solve the first $10$ or $20$ sentences, then use the lexical extender on the remaining data. 

%We employ the following procedure: First run the algorithm with a minimal number of categories---a start symbol, a non-lexical category, and a lexical category---and a large number of rules. The algorithm should find many solutions very easily, but of course they will not be very insightful solutions. Reduce the number of rules and repeat, until either the solutions become unique (up to permutation symmetries) or are no longer found. Use the solution(s) found with the smallest successful $n_r$ to generate new sentences.

\begin{table}[t]
    \begin{tabular}{l}
        $\textsf{S}\, \to\, \textsf{NP}_\mathsf{s}\, \textsf{VP}_\mathsf{s}\, |\, \textsf{NP}_\mathsf{p}\, \textsf{VP}_\mathsf{p}$\\
        $\textsf{NP}_\mathsf{s}\, \to\, \textsf{PN}\, |\, \textsf{D}_\mathsf{s}\, \textsf{N}_\mathsf{s}$ \\
        $\textsf{NP}_\mathsf{p}\, \to\, \textsf{N}_\mathsf{p}\, |\, \textsf{D}_\mathsf{p}\, \textsf{N}_\mathsf{p}$ \\
        $\textsf{VP}_\mathsf{s}\, \to\, \textsf{IV}_\mathsf{s,pres}\, |\, \textsf{IV}_\mathsf{past}\, |\, \textsf{TV}_\mathsf{s,pres}\, \textsf{NP}_\mathsf{s|p}\, |\, \textsf{TV}_\mathsf{past}\, \textsf{NP}_\mathsf{s|p}$ \\
        $\textsf{VP}_\mathsf{p}\, \to\, \textsf{IV}_\mathsf{p,pres}\, |\, \textsf{IV}_\mathsf{past}\, |\, \textsf{TV}_\mathsf{p,pres}\, \textsf{NP}_\mathsf{s|p}\, |\, \textsf{TV}_\mathsf{past}\, \textsf{NP}_\mathsf{s|p}$ \\
        $\textsf{D}_\mathsf{s}\, \to$ \textit{this $|$ every} \\
        $\textsf{D}_\mathsf{p}\, \to$ \textit{these $|$ all} \\
        $\textsf{N}_\mathsf{s}\, \to$ \textit{dog $|$ girl $|$ car $|$ child} \\
        $\textsf{N}_\mathsf{p}\, \to$ \textit{dogs $|$ girls $|$ cars $|$ children} \\
        $\textsf{PN}\, \to$ \textit{Kim $|$ Jody} \\
        $\textsf{IV}_\mathsf{s,pres}\, \to$ \textit{disappears $|$ walks} \\
        $\textsf{TV}_\mathsf{s,pres}\, \to$ \textit{sees $|$ likes} \\
        $\textsf{IV}_\mathsf{p,pres}\, \to$ \textit{disappear $|$ walk} \\
        $\textsf{TV}_\mathsf{p,pres}\, \to$ \textit{see $|$ like} \\
        $\textsf{IV}_\mathsf{past}\, \to$ \textit{disappeared $|$ walked} \\
        $\textsf{TV}_\mathsf{past}\, \to$ \textit{saw $|$ liked} \\
    \end{tabular}
    \caption{The rules of the second NLTK \cite{loper2002nltk} grammar, which makes a distinction between singular and plural as well as past and present tense.}
    \label{tab:feat0}
\end{table}

We will demonstrate the use of the category refiner and the lexical extender with our second toy grammar. Table \ref{tab:feat0} gives the rules of our second grammar. Unlike the first, this grammar contains categories with different tense and number, and a lexicon divided accordingly. 

\begin{table}[t]
    \centering
    \begin{tabular}{lll}
        $\textsf{S}$ & $\to$ & $\textsf{B}\,\textsf{A}$\\
        $\textsf{A}$ & $\to$ & $\textsf{D}\,\textsf{B}$ $|$ \textit{walks $|$ walk $|$ disappears $|$ disappear}\\
        $\textsf{B}$ & $\to$ & $\textsf{C}\,\textsf{B}$ $|$ \textit{dog $|$ cars $|$ Jody $|$ girl $|$ Kim} \\
            && $|$ \textit{dogs $|$ girls $|$ children $|$ car $|$ child}\\
        $\textsf{C}$ & $\to$ & \textit{this $|$ every $|$ these $|$ all}\\
        $\textsf{D}$ & $\to$ & \textit{likes $|$ see $|$ sees $|$ like}\\[3pt]
        \hline
         \\[-5pt]
        $\textsf{S}$ & $\to$ & $\textsf{B}_0\,\textsf{A}_0$ $|$ $\textsf{B}_1\,\textsf{A}_1$\\
        $\textsf{A}_0$ & $\to$ & $\textsf{D}_0\,\textsf{B}_0$ $|$ $\textsf{D}_0\,\textsf{B}_1$ $|$ \textit{walks $|$ disappears}\\
        $\textsf{A}_1$ & $\to$ & $\textsf{D}_1\,\textsf{B}_0$ $|$ $\textsf{D}_1\,\textsf{B}_1$ $|$ \textit{walk $|$ disappear}\\
        $\textsf{B}_0$ & $\to$ & $\textsf{C}_0\,\textsf{B}_2$ $|$ \textit{Jody $|$ Kim} \\
        $\textsf{B}_1$ & $\to$ & $\textsf{C}_1\,\textsf{B}_1$ $|$ \textit{cars $|$ dogs $|$ girls $|$ children}\\
        $\textsf{B}_2$ & $\to$ & \textit{dog $|$ girl $|$ car $|$ child}\\
        $\textsf{C}_0$ & $\to$ & \textit{this $|$ every}\\
        $\textsf{C}_1$ & $\to$ & \textit{these $|$ all}\\
        $\textsf{D}_0$ & $\to$ & \textit{likes $|$ sees}\\
        $\textsf{D}_1$ & $\to$ & \textit{see $|$ like}\\[3pt]
        \hline
         \\[-5pt]
        $\textsf{S}$ & $\to$ & $\textsf{B}_0\,\textsf{A}_0$ $|$ $\textsf{B}_1\,\textsf{A}_1$\\
        $\textsf{A}_0$ & $\to$ & $\textsf{D}_0\,\textsf{B}_0\,|\,\textsf{D}_0\,\textsf{B}_1\,|\,$\textit{walk$\,|\,$disappear$\,|\,$walked$\,|\,$disappeared}\\
        $\textsf{A}_1$ & $\to$ & $\textsf{D}_1\,\textsf{B}_0\,|\,\textsf{D}_1\,\textsf{B}_1\,|\,$\textit{walks$\,|\,$disappears$\,|\,$walked$\,|\,$disappeared}\\
        $\textsf{B}_0$ & $\to$ & $\textsf{C}_0\,\textsf{B}_2$ $|$ \textit{Jody $|$ Kim} \\
        $\textsf{B}_1$ & $\to$ & $\textsf{C}_1\,\textsf{B}_1$ $|$ \textit{cars $|$ dogs $|$ girls $|$ children}\\
        $\textsf{B}_2$ & $\to$ & \textit{dog $|$ girl $|$ car $|$ child}\\
        $\textsf{C}_0$ & $\to$ & \textit{this $|$ every}\\
        $\textsf{C}_1$ & $\to$ & \textit{these $|$ all}\\
        $\textsf{D}_0$ & $\to$ & \textit{likes $|$ sees}\\
        $\textsf{D}_1$ & $\to$ & \textit{see $|$ like}\\
    \end{tabular}
    \caption{\textit{Top}: A solution for the second toy grammar with $c=5$, $c_l=2$, $r_s=3$ for a dataset that had no past tense verbs. \textit{Middle}: The result of the refiner algorithm using the top solution as the starting point, but now with $f_\mathsf{S}=1$, $f_\mathsf{A}=f_\mathsf{C}=f_\mathsf{D}=2$, $f_\mathsf{B}=3$. \textit{Bottom}: The result of the extender algorithm using the middle solution as a starting point. We gave the extender an unabridged data set, including past tense verbs, and allowed $h=4$.}
    \label{tab:feat0sol}
\end{table}

First we use the main algorithm with an abridged data set that contains no past tense verbs. The settings that yield a unique solution are $c=5$, $c_l=2$, $r_s=3$, and $h=0$. {This turns out to be easier for our algorithm than the first grammar: In $100$ trials run on $100$ sentences, there were $83$ successes within $10^4$ iterations, for an average of $(2.7\pm0.4)\times10^3$ iterations per solution (using a metric parameter update rate of $10^{-2}$).} The solution is given in the top of Table \ref{tab:feat0sol}. An English-speaking reader will recognize that category \textsf{B} contains nouns, \textsf{C} contains determiners, and \textsf{A} and \textsf{D} contain, respectively, intransitive and transitive verbs. 

Next we pass this solution to the refiner. Once we choose the number of features for each category we reduce the allowed number of refined rules until the inferred solution becomes unique, just as with the main algorithm. One might suppose that $f_\mathsf{S}=1$ and $f_\mathsf{A}=f_\mathsf{B}=f_\mathsf{C}=f_\mathsf{D}=2$ would be a reasonable choice to accomplish splitting the categories into singular and plural, but in fact we need to increase $f_\mathsf{B}$ to $3$ in order to ensure uniqueness. The numbers of rules that work are $f_\mathsf{SBA}=f_\mathsf{BCB}=2$ and $f_\mathsf{ADB}=4$. The resulting solution is given in the middle of Table \ref{tab:feat0sol}. {In $100$ trials, all succeeded within $10^6$ iterations, for an average of $(1.5\pm0.2)\times10^4$ iterations per solution.}

Finally, we pass the refined solution to the extender, this time providing an unabridged data set (i.e., with past tense verbs) of $100$ sentences and allowing $h=4$. The unique solution is given in the bottom of Table \ref{tab:feat0sol}. The algorithm found this solution within $10^4$ iterations in $71/100$ trials, taking an average of $(6.6\pm0.9)\times10^3$ iterations per solution. With any lesser $h$ the algorithm fails to find any solutions within $10^4$ iterations. The only difference between this extended solution and the previous one is the addition of the four past tense verb forms which were absent from the abridged data set. Since the past tense forms are the same for the singular and plural cases, these four words require an allotment of four homographs, hence the necessity of $h=4$.

\section{Conclusions}\label{sec:conclusion}
This work constitutes a proof-of-concept for unsupervised grammar learning on a network with fully interpretable representations. We have illustrated how our algorithm can classify words into grammatical categories and infer context-free grammar rules to derive a given list of sentences. The user decides how fine-grained the grammar should be by choosing appropriate bounds on the numbers of categories and rules. If desired, one can feed the inferred grammar into a modified version of the algorithm to refine it further and capture features of each category, such as number or gender. One can also supply new sentences and ask the algorithm if they too are consistent with the inferred grammar and, if so, what the grammatical classifications are of any previously unseen words that appear in the new sentences.

Unlike gradient-descent approaches to grammar learning, in which the data sets are massive and the model can have billions of parameters that may not be easily interpretable, our model has only a handful of parameters --- of order $10^2$ bits for the syntax tensor in our experiments --- which are manifestly interpretable. Additionally, our algorithm only needs to see a few sentences before it begins to recognize the syntactic rules and successfully classifies the words into categories. 

We chose simple language fragments as our data sources in order to make it possible to definitively verify the success of our algorithm in reconstructing the grammar that generated the data. An obvious next step would be to try more complex language fragments (more words in the lexicon, more diverse syntax) and, eventually, natural language. Since there is no ``correct answer" for the grammar that generates natural language, testing on synthetic data is a necessary first step. 

We do not claim that our model is definitive and complete, but merely that it demonstrates a useful alternative approach to inferring grammar that is compact and interpretable. While our 1-hot category vector representations are `symbolic,' the computational architecture fits squarely in the connectionist framework. This melding of paradigms is made possible by the RRR algorithm \cite{elser2021learning} which is routinely used in problems with nonconvex constraints.

For readers expecting a leaderboard-style evaluation, we instead offer the following remarks. First, it is a remarkable fact that two radically different network architectures, giant ones with continuous parameters and (comparatively) tiny ones with discrete parameters, succeed at simultaneously solving three tasks without supervision: parsing sentences, discovering syntax rules, and assigning words to lexical categories. In the case of the small networks (this work) the evidence is direct, as we see from outputs such as Table \ref{tab:feat0sol}. For giant network methods (e.g. GPT-3 \cite{brown2020language}) the evidence is indirect but no less compelling because the language generated by them is highly grammatical. Giant-network/big-data methods are the clear choice for real-world applications, while the present approach seems better suited for answering questions such as: How many sentences ($10^1, 10^2, \ldots$) are needed to learn the concept of \textit{noun} (and the other parts of speech)? We would not expect a statistically trained giant network to find a continuation for ``Twas brillig, and the slithy toves ..." \cite{carroll2018jabberwocky}, while even nonsense data is fair game for our small networks. But the two approaches need not be mutually exclusive. After all, both employ distributed computing on network architectures and may just represent extreme points of a broader spectrum of methods.

\begin{acknowledgements}
We thank Jonathan Yedidia for helping us navigate the expansive field of computational linguistics.
\end{acknowledgements}

\appendix
\section{The ``divide" constraint projection}\label{sec:PA}

%A convenient way to impose the metric \eqref{metric} upon which the constraint projections are based is to define new $v$ variables that are scaled by $\mu$. In this appendix we work with these rescaled $v$ variables, where distance computations involving both $v$ and $t$ variables use an isotropic Euclidean metric. The parameter $\mu$ appears only in the discrete values: $v_\mathsf{X}\in\{0,\mu_\mathsf{X}\}$, $t\in\{0,1\}$.

Set $A$ is the set in which all $v$ and $t$ variables are discrete and every layer can be obtained from the layer below by applying a single rule. The variable copies allow us to treat each layer independently. For every sentence $s$, let $\Lambda_s$ be the length of the sentence. 

For each $\ell$ from $1$ to $\Lambda_s-1$, there are $\ell$ nodes at which one can apply the rule that transforms layer $\ell$ to layer $\ell+1$. Fixing the sentence $s$ and lower layer $\ell$, we use the following abbreviations, just in this section of the appendix, for the relevant variables:
\begin{align*}
v^{s \ell j \uparrow}\to &\;v^{j\uparrow}, \quad j=1,\ldots, \ell&\\
v^{s \ell+1 j \downarrow}\to &\; v^{j\downarrow},\quad j=1,\ldots, \ell+1&\\
t^{s \ell}\to &\; t.
\end{align*}
We recall that each $v$ is a category vector with possible subscripts $\mathsf{S}, \mathsf{A}, \mathsf{B}, \ldots$ while $t$ is an order-3 tensor with three such subscripts.

Before we begin any computations, we remark that the squared distance for projecting from an arbitrary $(v,t)$ to a point in set $A$ involves summands like \begin{equation*}\left(v^{j\uparrow}_\mathsf{X}\right)^2
\end{equation*}
if projecting to $0$, or
\begin{equation*}
\left(1-v^{j\uparrow}_\mathsf{X}\right)^2=1-2v^{j\uparrow}_\mathsf{X}+\left(v^{j\uparrow}_\mathsf{X}\right)^2
\end{equation*}
if projecting to $1$. The squared term on the right is present either way, so in comparing distances we need only consider the $1-2v^{j\uparrow}_\mathsf{X}$ term.

For all the nodes at which the rule is \textit{not} applied we must preserve the categories from this layer up to the next. For any node $j$ to the left of the rule application, preserving means that category vectors $v^{j\uparrow}$ and $v^{j \downarrow}$ should be equal, whereas for any node to the right it means $v^{j\uparrow}$ equals $v^{j+1 \downarrow}$. Projecting to the nearest pair of equal 1-hots for $j=1,\ldots,\ell-1$ means finding the category $\mathsf{L}$ that minimizes
\begin{equation*}
2\mu_\mathsf{L}^2\left(1- v^{j\uparrow}_\mathsf{L}-v^{j\downarrow}_\mathsf{L}\right).
\end{equation*}
Call this $\mathsf{L}(j)$ the ``left preservation category" for node $j$. Similarly, for every $j=2,\ldots,\ell$ we find the $\mathsf{R}$ that minimizes \begin{equation*}
2\mu_\mathsf{R}^2\left(1-v^{j\uparrow}_\mathsf{R}-v^{j+1\downarrow}_\mathsf{R}\right).
\end{equation*} 
Call $\mathsf{R}(j)$ the ``right preservation category" for node $j$. 

{For each of the possible rule application positions $i=1, \ldots, \ell$, we compute $d^i$, the squared distance associated with preserving categories when applying the rule at position $i$:
\begin{itemize}
    \item For every node $j=1,\ldots,i-1$ to the left of the rule application we add to $d^i$ the distance for projecting $v^{j\uparrow}$ and $v^{j\downarrow}$ to $\mathsf{L}(j)$. 
    \item For $j=i+1,\ldots,\ell$ we add the distance for projecting $v^{j\uparrow}$ and $v^{j+1\downarrow}$ to $\mathsf{R}(j)$.
\end{itemize}
Note that $d^i$ does not depend on which rule is applied, only on the position $i$ where it would be applied.

As for the syntax tensor, we can handle its computations without knowing the position at which the rule is to be applied. We know that the tensor must contain at most $r_s$ rules, that every \textsf{YZ} can have at most one \textsf{X} such that $t_\mathsf{XYZ}=1$, and that the rules are subject to the restrictions in section \ref{sec:problem}. So for every \textsf{YZ} with \textsf{Y}$\neq$\textsf{Z} we find the $\mathsf{X}_\mathsf{YZ}$ (subject to the restrictions) that maximizes $t_{\mathsf{X}_\mathsf{YZ}\mathsf{YZ}}$. We rank the pairs \textsf{YZ} according to the value of $t_{\mathsf{X}_\mathsf{YZ}\mathsf{YZ}}$ and for the $r_s$ pairs with the largest $t_{\mathsf{X}_\mathsf{YZ}\mathsf{YZ}}$ we set \begin{equation*} 
    \left[P_A\left(t\right)\right]_{\mathsf{XYZ}}=\begin{cases}1&\text{if}\;\mathsf{X}=\mathsf{X}_\mathsf{YZ}\;\text{and}\;t_{\mathsf{X}_\mathsf{YZ}\mathsf{YZ}}\geq\frac12\\
    0&\text{otherwise.}\end{cases}.
\end{equation*} 
For all other \textsf{YZ}, we set $\left[P_A\left(t\right)\right]_{\textsf{XYZ}}=0$.

Next, we loop through all the allowed rules \textsf{X} $\to$ \textsf{YZ} and compute the extra distance $d_\mathsf{XYZ}$ that would be required to accommodate \textsf{X} $\to$ \textsf{YZ} in the syntax tensor. Just as with the category vectors, we only need to consider terms of the form $1-2t_\mathsf{XYZ}$, and only for the elements of the tensor that are affected by our choice of which syntactic rule to use. 
\begin{itemize}
    \item If \textsf{YZ} is one of the $r_s$ chosen pairs, then we need to ensure we do not have two rules with the same \textsf{YZ}. If $t_{\mathsf{X}_\mathsf{YZ}\mathsf{YZ}}\geq1/2$, we remove $\mathsf{X}_\mathsf{YZ}\to\mathsf{YZ}$ from the rule set and add $\mathsf{X}\to\mathsf{YZ}$, which involves a distance of
    \begin{equation}
        d_\mathsf{XYZ}=(1-2t_\mathsf{XYZ})-(1-2t_{\mathsf{X}_\mathsf{YZ}\mathsf{YZ}})=2(t_{\mathsf{X}_\mathsf{YZ}\mathsf{YZ}}-t_\mathsf{XYZ}).
    \end{equation}
    If $t_{\mathsf{X}_\mathsf{YZ}\mathsf{YZ}}<1/2$, then $\mathsf{X}_\mathsf{YZ}\to\mathsf{YZ}$ is not in the rule set to begin with, so the distance is just $1-2t_\mathsf{XYZ}$.
    \item If \textsf{YZ} is not one of the chosen pairs, then we need to make sure we do not have more than $r_s$ pairs with a syntactic rule. So we look at the chosen pair that had the $r_s$-th greatest $t_{\mathsf{X}_\mathsf{YZ}\mathsf{YZ}}$ --- let us call it $t_r$ for short. The distance is 
    \begin{equation}
        d_\mathsf{XYZ}=\begin{cases}2(t_r-t_\mathsf{XYZ})&t_r\geq\frac12\\
    1-2t_\mathsf{XYZ}&t_r<\frac12\end{cases}.
    \end{equation}
\end{itemize}

Putting everything together, the squared distance required to use \textsf{X} $\to$ \textsf{YZ} at position $i$ is
\begin{multline}
    d^i_\mathsf{XYZ}=\mu_\mathsf{X}^2\left(1-2v^{i\uparrow}_\mathsf{X}\right)+\mu_\mathsf{Y}^2\left(1-2v^{i\downarrow}_\mathsf{Y}\right)\\
    +\mu_\mathsf{Z}^2\left(1-2v^{i+1\downarrow}_\mathsf{Z}\right)+d^i+d_\mathsf{XYZ}.
\end{multline}
After finding the \textsf{XYZ} and $i$ that minimize this distance, we set the preservation categories:
\begin{equation}
    \left[P_A\left(v^{j\uparrow}\right)\right]_{\mathsf{L}(j)}=1=\left[P_A\left(v^{j\downarrow}\right)\right]_{\mathsf{L}(j)}
    \label{eq:Lpres}
\end{equation}
for $j=1,\ldots,i-1$ and
\begin{equation}
    \left[P_A\left(v^{j\uparrow}\right)\right]_{\mathsf{R}(j)}=1=\left[P_A\left(v^{j+1\downarrow}\right)\right]_{\mathsf{R}(j)}
    \label{eq:Rpres}
\end{equation}
for $j=i+1,\ldots,\ell$. We then set the three category vectors involved in the syntactic rule:
\begin{equation*}\label{eq:ruleapp}
    \left[P_A\left(v^{i\uparrow}\right)\right]_\mathsf{X}=\left[P_A\left(v^{i\downarrow}\right)\right]_\mathsf{Y}=\left[P_A\left(v^{i+1\downarrow}\right)\right]_\mathsf{Z}=1.
\end{equation*}}
We set all other components of the category vectors to $0$, with one exception: If $h>0$, then for the vectors in the top layer we simply round those components to $0$ or $1$, whichever is nearer. The purpose of this exception is to allow for homographs, as the top layer category vectors imply the lexical rules.

{Finally, the syntax tensor: If \textsf{YZ} was one of the $r_s$ chosen pairs and $\mathsf{X}\neq\mathsf{X}_\mathsf{YZ}$, then we set 
\begin{equation*} 
    \left[P_A\left(t\right)\right]_{\mathsf{X}_\mathsf{YZ}\mathsf{YZ}}=0,\qquad\left[P_A\left(t\right)\right]_\mathsf{XYZ}=1.
\end{equation*}
If \textsf{YZ} was not one of the chosen pairs, then we go back once again to the the $r_s$-th greatest of the chosen pairs (the one that gave us $t_r$) and set it to $0$, then set $\left[P_A\left(t\right)\right]_\mathsf{XYZ}=1$.
}

\section{The ``concur" constraint projection}\label{sec:PB}
Set $B$ is the set in which all the copies of the syntax tensor agree, the $\uparrow$ and $\downarrow$ copies of the category vectors agree, and all top-layer category vectors for a given word in the lexicon also agree. To ensure that the copies of the syntax tensor agree we set
\begin{equation}
    P_B\left(t^{s\ell}\right)=\frac{\sum_{s'\ell'}t^{s'\ell'}}{\sum_{s'\ell'}}
\end{equation}
for all $s$ and $\ell$. %We must also limit the number of syntactic rules, so if \begin{equation}
%r=\sum_{\mathsf{XYZ}}\left(\bar{t}_{\mathsf{XYZ}}\right)^2 > r_s,
%\end{equation}
%we also rescale every element of $\bar{t}$ by the factor $\sqrt{r_s/r}$. This is the distance-minimizing way to ensure that all copies of the rule tensor agree and that the sum of squares of the elements of the rule tensor does not exceed $r_s$.

Next, we ensure that the copies of the category vectors agree by setting
\begin{equation}
    P_B\left(v^{s\ell i\uparrow}\right)=\frac{v^{s\ell i\uparrow}+v^{s\ell i\downarrow}}{2}=P_B\left(v^{sli\downarrow}\right)
\end{equation}
for all $s$, $l$, and $i$ except in the bottom and top layers. To ensure that all categories are used at least once, for every \textsf{X} that is not the start symbol and is not one of the $c_l$ designated lexical categories, we keep track of the $(s,\ell,i)$ with the largest value of $[P_B(v^{s\ell i\uparrow})]_\mathsf{X}$. If this largest value is less than $1$ we set
\begin{equation}
    [P_B(v^{s\ell i\uparrow})]_\mathsf{X}=1=[P_B(v^{s\ell i\downarrow})]_\mathsf{X}
\end{equation}
at the $(s,\ell,i)$ at which the largest value occurred.

In the bottom layer there is no $\downarrow$ copy, so $P_B(v^{s00\uparrow})=v^{s00\uparrow}$. In the top layer there are no $\uparrow$ copies but here we must enforce the lexical rule restrictions. Let $\Pi_w$ be the set of ordered pairs $(s,i)$ specifying the sentences $s$ and positions $i$ within the sentence at which $w$ appears. For each $w\in\Omega$ we set 
\begin{equation}
    P_B\left(v^{s\Lambda_s i\downarrow}\right)=\bar{v}^w=\frac{\sum_{(s',i')\in\Pi_w} v^{s'\Lambda_{s'}i'\downarrow}}{|\Pi_w|}
\end{equation}
for all $(s,i)\in\Pi_w$. We need to ensure that all of the designated lexical categories are used here, so for each of these categories \textsf{X} we record $\bar{v}^w_\mathsf{X}$ for every word $w$. If there is no $w$ such that $\bar{v}^w_\mathsf{X}\geq1$, then in principle the distance-minimizing change would be to choose the $w^*$ such that $$|\Pi_{w^*}|\left(1-\bar{v}^{w^*}_\mathsf{X}\right)^2$$ is smallest and set
\begin{equation}
    [P_B(v^{s\Lambda_s i\downarrow})]_\mathsf{X}=1
\end{equation}
for all $(s,i)\in\Pi_{w^*}$. In practice, we have discovered that the algorithm works even more efficiently if we instead choose $w^*$ such that $$\left(1-\bar{v}^{w^*}_\mathsf{X}\right)^2$$ is smallest. 

Just like the upper bound on the number of syntactic rules, the upper bound on the number of lexical rules is imposed via the L2 norm. Including the $h>0$ homograph allowance, we must check if $$\sum_w \left(\bar{v}^w\right)^2=r\leq|\Omega|+h.$$ If not, we multiply the top layer category vectors by $$\sqrt{(|\Omega|+h)/r}.$$ When $h=0$ this rescaling is not necessary because the discrete $A$ constraint already ensures the category vector for each word is 1-hot.

\section{Divide-and-concur networks in the language of feed-forward networks}\label{sec:DC}

This appendix is aimed at the 99.9999\% of readers who are familiar with feed-forward neural networks but have never encountered divide-and-concur (DC), let alone its deployment on networks. The treatment is light and relies on the power of language and analogy to describe the unfamiliar in familiar terms.

The ``$A$ constraint" of DC comes closest to the non-linear \textit{activation functions} of feed-forward networks. Consider the extreme case of \textit{step-activation}. If the inputs to the activation functions are approximately two-valued, say 0/1 (as a result of other step activations), then with suitable \textit{bias parameter} the activation function can model \textsc{Or} gates, \textsc{And} gates, and things in between (depending on the number of inputs). We will consider the simplest case where all the activation functions/gates in the network have two inputs and the bias decides whether each is to be an \textsc{Or} or \textsc{And}.

\begin{figure}[b]
    \centering
    \includegraphics[width=.35\textwidth]{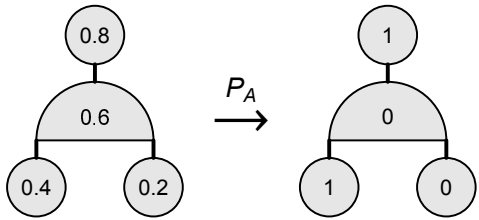}
    \caption{Logic assignment projection, $P_A$, applied to a gate (semicircle) taking two inputs. The gate may have two states: $f=0$ for \textsc{Or} and $f=1$ for \textsc{And}.}
    \label{fig:gateproj}
\end{figure}

Figure \ref{fig:gateproj} shows how two-input step-activation functions would be implemented in a DC network. Let the variables for the two inputs be $x_1$ and $x_2$ and the output be $y$. Instead of a bias parameter, the ``state" of the gate is encoded by a 0/1-valued parameter $f$. By convention $f=0$ is an \textsc{Or} gate and $f=1$ is an \textsc{And} gate. On the left of the figure we see the network with values $x_1=0.4$, $x_2=0.2$, $y=0.8$ and $f=0.6$.

Here is where we see some important differences between DC and feed-forward networks. First, there is no ``feed-forward" at all. Instead there is a constraint that acts equally between inputs, the output, and the state $f$ of the gate. This constraint is imposed by the operator $P_A$ (projection to the $A$ constraint) that appears through its reflection $R_A$ in the update equation \eqref{eq:RRR}. The result of $P_A$ is shown in the right of the figure and is simply the 0/1 assignment to $(x_1,x_2,y,f)$ that is (i) consistent with the \textsc{Or}/\textsc{And} interpretation of $f$ (eight possible assignments) and (ii) minimizes the distance
\begin{equation}
(x_1-0.4)^2+(x_2-0.2)^2+(y-0.8)^2+(f-0.6)^2.
\end{equation}
The reader is invited to check that the other seven valid assignments have a greater distance.
 
The $P_A$ computation just described is performed synchronously on all the gates of the network. This is possible because each gate output, such as $y$, and the gate inputs it feeds into, say $x_3, x_4, \ldots$ , are allocated different variables. This is the origin of the term ``divide" in DC: The constraints of the problem are divided into independent sets. To recover a solution to the original problem the other projection operator, $P_B$, imposes equality of the variable copies. In this case
\begin{equation}
y=x_3=x_4=\cdots
\end{equation}
The ``concur value" $c$, shared by all these variables, minimizes
\begin{equation}
(c-y)^2+(c-x_3)^2+(c-x_4)^2+\cdots ,
\end{equation}
and is equal to the average of the numbers $y, x_3, x_4, \ldots$

Depending on the application, the concur values may be supplanted by known values, say at the inputs and outputs of the network in the case of \textit{supervised learning}, or, in the case of \textit{unsupervised learning}, just at the outputs when only the outputs are known (and the network is also tasked with reconstructing an input that goes with each output).

Because the DC \textit{learning algorithm} or \textit{optimizer} is built from the operators $P_A$ and $P_B$ just described, we see that there are no gradient computations or calculus of any kind. Instead, the RRR update \eqref{eq:RRR} generated by $P_A$ and $P_B$ is applied over and over until there is a fixed point. Writing the update in terms of the projections (instead of the reflectors),
\begin{equation}
x'=x+\beta\left(P_B(2P_A(x)-x)-P_A(x)\right) ,
\end{equation}
we see that $x'=x$ implies that
\begin{equation}
P_B(2P_A(x)-x)=P_A(x)=x_\mathrm{sol}
\end{equation}
is a solution because it is a point that lies in both constraint sets, $A$ and $B$. In our example, all the gate inputs, outputs and states will be 0/1 (set $A$) and the output of each gate will agree with the input it supplies to other gates in the network or a known output value (set $B$). Notice that the simpler update $x'=P_B(P_A(x))$ does not have this property. If $x'=x$, then it is possible that $P_A(x)\ne x$ and therefore does not lie in set $B$ (violating concur). In fact, because the set $A$ is \textit{nonconvex}, this scenario is in practice highly probable and makes ``alternating projections" not a viable update rule.

The synchrony of the $P_A$ and $P_B$ operations represents another difference with feed-forward networks. Training the latter involves passing information forward in the \textit{inference} part of the update and then backward when \textit{back-propagating} the gradient information. By contrast, in DC information is propagated (via the action of $P_A$ and $P_B$) in both directions in each application of the update rule. The \textit{hyperparameter} $\beta$ controls the rate at which this information is propagated, and is roughly analogous to the \textit{learning rate} hyperparameter $\eta$ of feed-forward networks. Gradient descent is only exact in the limit $\eta\to 0$, but any $\beta\in(0,2)$ gives local convergence to fixed points of the RRR update (Theorem 26.11 of Bauschke et al. in \cite{bauschke2011convex}).

Whereas a large value, say $\beta=1$, makes sense because the variables see significant change at a higher rate, there is also a good reason to keep $\beta$ small. Because information propagates at a finite rate, only between connected gates (neurons) in each update, keeping $\beta$ small ensures there is more time for the information to find its way around the entire network before variables are significantly changed. This is a good strategy when networks are small and learning \textit{representations} needs to be more of a cooperative process than is suggested by the \textit{lottery ticket hypothesis} \cite{frankle2018lottery}.

When DC is applied to training networks, new hyperparameters naturally arise. Notice that in our example of a network of gates the state $f$ of a gate was treated no differently from the node variables (gate inputs/outputs) by the $P_A$ operator. In retrospect, it seems arbitrary that the change in $f$ was given the same weight as the changes at the nodes. By introducing a multiplier $\mu>0$ to the term in the distance for $f$ one can make the gates more ($\mu<1$) or less ($\mu>1$) compliant than the nodes when projecting to the logic assignment. This is important in that it provides an intervention for one of DC's failure modes. This is when one type of variable, say the node variables in our example, remain essentially static and only the other type, the $f$'s of the gates, are changing significantly. If the former variables are stuck on the wrong values, the constraint problem for the latter is insoluble and the algorithm executes a fruitless search. To remedy this one increases $\mu$, making the gate states less compliant, thereby forcing the node variable to try other assignments. Conversely, when only the node variables are changing, and the gate-state variables are stuck on the wrong values, $\mu$ should be decreased.

\textit{Weight-sharing} is important in many applications and is another instance where feed-forward networks and DC differ. The best known example is \textit{convolutional networks}, where the translational symmetry of feature detection in images is exploited by allocating a single set of weight parameters to all the neurons in the lowest layers of the network. Similarly, in the grammar inference problem there should be a single set of weights that define the syntax rules wherever they are applied when parsing a sentence.

DC handles the sharing of variables differently. First, note that we use the term ``variables" even for the parameters (e.g. weights) that are learned. We do this because DC trains on \textit{data-batches} synchronously. In the example above, of learning \textsc{Or}/\textsc{And} assignments to gates that explain all the data in a batch (pairs of network inputs/outputs), there would be different node variables for each network instantiation while the gate ``parameter-variables" ($f$'s) are shared across the batch. In keeping with the local mindset, DC allocates different gate parameter-variables to the different data instantiations of the network (divide) and uses a concur constraint to enforce equality (sharing) of those parameter-variables. The shared parameter-variables in the grammar inference network of the main text are the elements of the syntax tensor $t$. These are shared, via the DC trick, across all sentences and layers of the parse trees where syntax rules are applied.

\textit{Batch normalization} is the most global update rule in the training of feed-forward networks. It too has a counterpart in DC networks when there are global constraints, such as the upper bounds on the number of syntactic and lexical rules in the grammar (items 2 and 3 in section \ref{sec:restrictions}). These are implemented by rescaling the concur values (appendix \ref{sec:PB}).

\bibliographystyle{unsrt}
\bibliography{grammar}

\end{document}